\documentclass[letterpaper, 10 pt, conference]{ieeeconf}
\IEEEoverridecommandlockouts
\usepackage{cite}
\usepackage{amsmath,amssymb,amsfonts}
\usepackage{algorithmic}
\usepackage{graphicx}
\usepackage{textcomp}
\usepackage{xcolor}
\usepackage{stfloats}
\usepackage{url}

\begin{document}

\title{Impact of a Soft Wearable Back-Support Device on Postural Stability
during Trip-Like Perturbations\\
%{\footnotesize \textsuperscript{*}Note: Sub-titles are not captured in Xplore and should not be used}
\thanks{The authors are with the School for Engineering of Matter, Transport and Energy, Arizona State University, Tempe, AZ, USA 
\\\textsuperscript{*}Corresponding author: Hyunglae Lee (hyunglae.lee@asu.edu).}
}

%\author{
%\IEEEauthorblockN{
%\begin{tabular}{ccc}
%Yuanhao Chen & Rohan Khatavkar & Soubhagya Nayak \\
%\textit{Ira A. Fulton Schools of Engineering} & \textit{Ira A. Fulton Schools of Engineering} & \textit{Ira A. Fulton Schools of Engineering} \\
%\textit{Arizona State University} & \textit{Arizona State University} & \textit{Arizona State University} \\
%Tempe, USA & Tempe, USA & Tempe, USA \\
%\end{tabular}
%}
%\\[1.5ex]
%\IEEEauthorblockN{
%\begin{tabular}{cc}
%Jiefeng Sun & Hyunglae Lee\textsuperscript{*} \\
%\textit{Ira A. Fulton Schools of Engineering} & \textit{Ira A. Fulton Schools of Engineering} \\
%\textit{Arizona State University} & \textit{Arizona State University} \\
%Tempe, USA & Tempe, USA \\
%\end{tabular}
%}
%}

\author{Yuanhao Chen, Rohan Khatavkar, Soubhagya Nayak, \\ Jiefeng Sun, and Hyunglae Lee\textsuperscript{*}}

\maketitle

\begin{abstract}
The effectiveness of a soft wearable back-support device in enhancing postural stability was investigated under trip-like perturbations using two experimental paradigms: perturbed standing and perturbed walking. Healthy subjects completed trials under three different back-support conditions: no device, device worn with low stiffness, and device activated with high stiffness. Whole-body stability was quantified using the minimum Margin of Stability (MOS) at the point of maximal instability. Results demonstrated increased MOS during device use, indicating enhanced postural stability. In standing, MOS increased significantly with device stiffness, whereas in walking, both device conditions improved MOS relative to no device but did not differ significantly from each other. These findings highlight the potential of soft wearable back-support devices with adjustable stiffness to improve reactive balance control against external perturbations, with important implications for fall prevention. Future research should explore personalized stiffness optimization and evaluate efficacy in populations at elevated risk of falls. 

\end{abstract}

%\begin{IEEEkeywords}

%wearable device, back support, soft back exosuit, gait stability, margin of stability, balance control
%\end{IEEEkeywords}

\section{Introduction}
Falls among older adults are a major health concern due to their high incidence and severe consequences. Approximately one in four community-dwelling seniors experience a fall each year, often resulting in serious injuries such as hip fractures or head trauma \cite{Diaz2022}. Slips and trips are among the most common precipitating causes, accounting for roughly 60\% of outdoor falls in adults over 70 \cite{Pai2014}. These events highlight the importance of developing effective strategies for improving postural control and stability under unexpected perturbations.

Devices that assist hip flexion or prevent excessive trunk flexion may improve balance. Walkers and canes are the simplest devices that stop excessive trunk flexion. Nevertheless, in order to respond to a perturbation, the user must actively move these devices. These devices might not be efficacious because older adults have diminished reflexes \cite{Gell2015}. Exoskeletons that offer assistive torque at the hip may enhance stability autonomously. Preliminary studies with hip exoskeletons have shown their potential to improve stability under perturbations \cite{tagliaferri2026systematic,leestma2024dynamic}. However, hip exoskeletons have an indirect impact on upper-body dynamics and trunk stabilization. Trunk motion plays a crucial role in fall prevention, as decreased trunk stability has been associated with increased fall risk among older adults \cite{Nevisipour2023,Vanderlinden2024}. Back support devices (BSDs) that provide assistive trunk torque may be more efficacious in enhancing stability than lower-limb exoskeletons. However, commercially available passive BSDs like Laevo FLEX (\url{en.laevo.nl}) and SuitX BackX (\url{www.suitx.com}) do not significantly improve postural stability in quiet stance \cite{park2021effects}. These exoskeletons have also shown adverse effects on stability during level walking, single-step recovery following a forward loss of balance, and slip-like and trip-like perturbations on a treadmill \cite{park2022effects, park2024wearing, dooley2024occupational}. These studies attribute the adverse or insignificant impact on stability to the added 2.8 - 4.3 kg weight of the BSDs. Recent studies have shown that lightweight non-extensible lumbar belts can improve dynamic postural stability in healthy individuals, presumably by increasing intra-abdominal pressure and spinal stiffness \cite{Bai2023, Azadinia2019}. Thus, a lightweight BSD that increases trunk stiffness may have the potential to improve stability.

%Despite growing interest, few studies have examined explicitly whether wearing a back-support device helps or hinders a person’s ability to recover stability after a trip \cite{Azadinia2019}. While this and similar robotic designs have demonstrated promising potential in supporting posture and enhancing mobility, the specific effects of such devices on whole-body stability, particularly under dynamic perturbation conditions like trips and slips, remain inadequately understood \cite{Li2023}.

The present study investigates how our previously developed lightweight (1.2 kg) BSD \cite{Khatavkar2025} with tunable stiffness affects postural stability during controlled perturbation experiments. The BSD was tested under two experimental conditions: trip-like perturbation during standing and walking. For controlled evaluation of postural stability in laboratory settings, perturbations can be introduced using unexpected obstacles, low-friction sliding platforms, sudden changes in treadmill speed, or waist-pull devices to mimic real-life trips \cite{Diaz2022}. In this study, we simulate trip-like perturbations on a split-belt treadmill with sudden changes in velocity. We first hypothesize that the use of our lightweight BSD will improve stability by increasing trunk stiffness during standing and walking trips. We also hypothesize that increasing the stiffness of our BSD will further improve stability. %By evaluating the margin of stability (MOS) responses across different stiffness levels of the BSD, this study provides new insights into the potential of lightweight back-support technology for fall prevention.

\section{Methods}

\subsection{Subjects}
Five young healthy adult males (age: 23 ± 2 years; weight: 79.2 ± 6.8 kg; height: 1.81 ± 0.08 m) participated in this study. All subjects had no history of neurological disorders, musculoskeletal injuries, or impairments affecting the lower extremities or back. The study protocol was approved by the Institutional Review Board of Arizona State University (STUDY00020698), and written informed consent was obtained from all subjects prior to data collection.

\subsection{Soft Wearable Back-Support Device}

\begin{figure}[htbp]
    \centering
    \includegraphics[width=1\columnwidth]{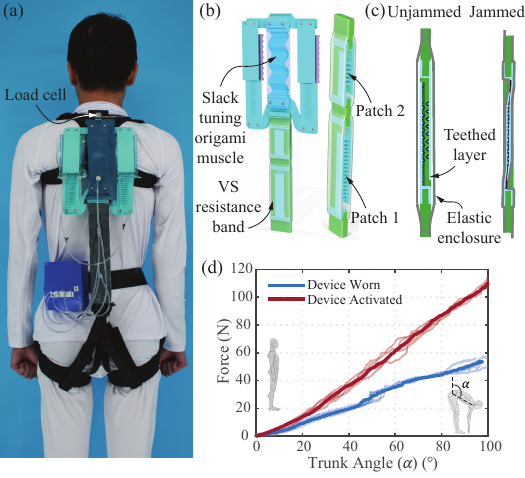}
    \caption{Device overview and characterization. (a) Subject wearing the device during experimental setup. (b), (c) Main components of the device, including the variable-stiffness resistance band and jamming status. (d) Measured device force as a function of trunk flexion angle, illustrating the device’s elastic response.}
    \label{fig:device}
\end{figure}

The soft wearable BSD used in this study is designed to assist trunk extension by generating a tensile force between the upper back and the thighs (Fig.~\ref{fig:device}(a)). As the trunk flexes forward, the relative displacement between these attachment points increases, elongating the elastic element and generating a restoring force that assists trunk extension. The device consists of two primary functional modules connected in series (Fig.~\ref{fig:device}(b)): (1) a variable-stiffness (VS) resistance band, and (2) a slack-tuning origami muscle assembly.

The VS resistance band serves as the primary elastic element responsible for generating the assistive force. Its stiffness is tuned using two teethed layer-jamming patches. Each airtight patch encloses two teethed layers (Fig.~\ref{fig:device}(b)). When vacuum pressure is applied to a patch (jammed state), the teethed layers engage, making the patch inextensible (Fig.~\ref{fig:device}(c)). Thus, the active length of the resistance band decreases, thereby increasing its stiffness. When the vacuum is released (unjammed state), the teethed layers slide freely.

In this study, two stiffness conditions were tested: (i) a low-stiffness state with both patches unjammed, and (ii) a high-stiffness state with both patches jammed. These two states correspond to the \textit{Device Worn} and \textit{Device Activated} conditions, respectively. The measured device force as a function of trunk flexion angle for each stiffness condition is shown in Fig.~\ref{fig:device}(d), demonstrating an increase in force output across the range of motion under the high stiffness condition.

Slack tuning is achieved using a compact origami muscle connected in series with the VS resistance band (Fig.~\ref{fig:device}(b)). By contracting or relaxing this origami muscle, the user can define the device's slack length. %At the maximum contraction of the origami muscle, an initial pretension of 20 N can be applied.

As shown in Fig.~\ref{fig:device}(d), the device force in the two stiffness conditions is indistinguishable up to 5 N due to an initial compliant interaction between the human body and the device. Thus, we used the origami muscle assembly to apply an initial pretension of 5 N in the upright posture for standing with perturbation trials. For walking with perturbation trials, we set the pretension to 0~N to avoid restriction during walking.

\subsection{Experimental Setup}
Experimental data were collected using a dual-belt treadmill with embedded force plates (Bertec Corporation, OH, USA), which recorded ground reaction forces at 2000 Hz. Whole-body kinematics were captured with an eight-camera motion capture system (Vicon Motion Systems Ltd., Oxford, UK) at 200 Hz, using a simplified Plug-in Gait marker set. Device force was measured with a load cell (Model LCM300, FUTEK, USA) attached between the resistance band and thigh connection point. 

All hardware systems were synchronized through a real-time controller (Speedgoat, MA, USA). Subjects wore tight-fitting clothing (ASICS Inc., Kobe, Japan) to minimize marker displacement, and a safety harness (Bertec Corporation, OH, USA) was used to prevent falls during treadmill perturbations.

\subsection{Experimental Protocol}

Subjects completed two experimental conditions: Standing with Perturbations (SWP) and Walking with Perturbations (WWP), both designed to evaluate the effect of our BSD on postural stability during reactive balance responses to trip-like perturbations. Each subject underwent trials under three device conditions: (1) No Device, (2) Device Worn, corresponding to the low-stiffness state with both patches unjammed, and (3) Device Activated, corresponding to the high-stiffness state with both patches jammed.

\begin{figure}[htbp]
    \centering
    \includegraphics[width=1\columnwidth]{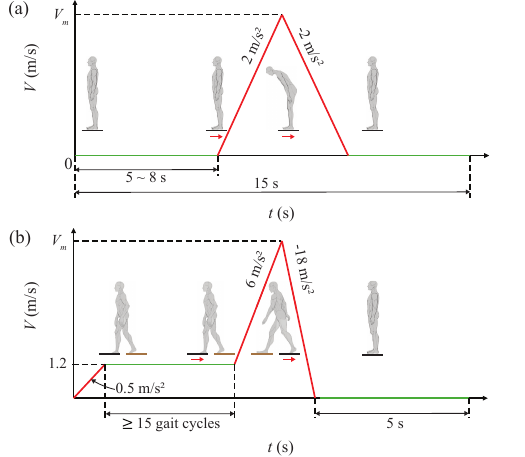}
    \caption{Treadmill velocity variations and corresponding subject postural changes during each trial. (a) SWP experiment. (b) WWP experiment.}
    \label{fig:protocol}
\end{figure}

In the SWP task, subjects stood on a dual-belt treadmill with feet shoulder-width apart and knees fully extended. Perturbations were induced by a posterior belt acceleration of 2 m/s². Subjects were instructed not to move their feet and to allow natural trunk lean. The perturbation velocity ($V_m$) was individually calibrated to elicit trunk flexion of 35 ± 3° (Fig.  \ref{fig:protocol}(a)). Each subject completed 60 trials: 20 without the device and 40 with the device (randomized between the two stiffness levels). Rest breaks were provided every 20 trials.

In the WWP task, subjects walked at 1.2 m/s, and perturbations were triggered at dominant-leg toe-off via rapid belt acceleration (6 m/s²) followed by deceleration (–18 m/s²). The perturbation velocity ($V_n$) was individually tuned to induce 30 ± 3° trunk flexion (Fig. \ref{fig:protocol}(b)). Each trial consisted of five perturbations (once every 15 gait cycles). Subjects completed 12 trials: four control (No device) and eight experimental trials across the two stiffness levels (Device Worn and Device Activated). Rest was provided every four trials.

%To ensure consistent device engagement across tasks, initial interaction forces were empirically set to 15 N (SWP) and 10 N (WWP), adjusted through belt length tuning to offset the device’s nonlinear response. Trial and block orders were randomized to minimize learning and fatigue effects.

\subsection{Data Processing and Statistical Analysis}
\begin{figure}[htbp]
    %\centering
    \includegraphics[width=1\columnwidth]{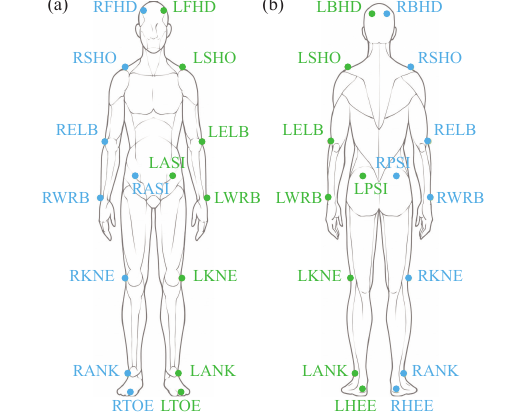}
    \caption{The 22-marker simplified Full Body Plug-in Gait Model.}
    \label{fig:markerset}
\end{figure}

%1. Simplified marker set to estimate COM
In both the SWP and WWP experiments, a simplified Full Body Plug-in Gait model (Vicon Motion Systems Ltd., Oxford, UK) was implemented to estimate whole-body center of mass (COM) positions. The protocol utilized 22 reflective markers strategically placed on the human body to define several body segments, balancing measurement accuracy with practical efficiency. Body segment mass proportions and approximate COM locations followed standardized biomechanical parameters, as detailed in the Vicon help documentation \cite{Vicon2023}.
Unlike the full-body Plug-in Gait marker set \cite{Vicon2023}, this simplified version included only the markers shown in Fig. \ref{fig:markerset}. Other markers from the full-body Plug-in Gait marker set that interfered with the BSD were excluded.

%The 4-point central finite difference method demonstrated superior performance over conventional first-order derivative approaches for calculating COM velocity. This method provides enhanced derivative approximation accuracy by reducing truncation errors while simultaneously mitigating the effects of signal noise \cite{LeVeque1998}.

The margin of stability (MOS) provides a direct mechanical measure of balance stability by quantifying the distance between the extrapolated center-of-mass (X$_{COM}$) and the boundary of support (BOS), thereby indicating how close a person is to a falling threshold \cite{Inagaki2023}. This makes MOS highly sensitive to dynamic stability during perturbations, as it captures whether a disturbance drives the COM beyond stability limits or whether sufficient margin remains for recovery \cite{Bruijn2013}. 

During the SWP trials, subjects maintained a parallel stance with shoulder-width foot placement and were restricted from active foot repositioning during perturbations. To quantify anterior stability, the boundary of support in the AP direction ($BOS_A$) was defined as the average coordinate value between the left and right toe markers (LTOE and RTOE) along the AP direction. 

%\begin{figure}[htbp]
    %\centering
    %\includegraphics[width=0.6\columnwidth]{fig/SWP.png}
    %\caption{The anterior boundary of support in SWP experiment.}
    %\label{fig:SWP BOS}
%\end{figure}

\begin{figure}[htbp]
    \centering
    \includegraphics[width=1\columnwidth]{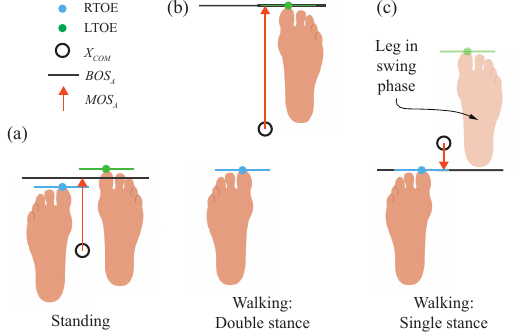}
    \caption{The anterior boundary of support and the anterior margin of stability in SWP and in WWP experiments.}
    \label{fig:WWP BOS}
\end{figure}

During the WWP trials, the $BOS_A$ dynamically transitioned across gait phases. As illustrated in Fig. \ref{fig:WWP BOS}(b), during double stance phases (both feet grounded), $BOS_A$ was determined by the AP coordinate of the leading foot's toe marker. In single stance phases (one-foot swing), $BOS_A$ corresponded exclusively to the AP coordinate of the weight-bearing foot's toe marker (Fig. \ref{fig:WWP BOS}(c)), where the transparent foot indicates the swing foot. 

% -----------------------------------
%2. Equations to calculate MOS
In both SWP and WWP experiments, perturbations were designed exclusively to induce forward-directed balance loss through trip-like disturbances. Consequently, MOS was analyzed solely in the anterior direction. X$_{COM}$ was calculated using Equation (\ref{eq:XCOM}) \cite{Hof2005}:

\begin{equation}
X_{\text{COM}} = P_{\text{COM}} + \frac{V_{\text{COM}}}{\sqrt{g/l}}
\label{eq:XCOM}
\end{equation}
where \textit{$P_{COM}$} and \textit{$V_{COM}$} were the horizontal position and velocity of the COM in AP direction, respectively. \textit{g} was the gravity constant (9.81 m/$s^2$), and \textit{l} was subject leg length multiplied by 1.2 \cite{Hof2005}. Leg length was calculated as the vertical distance between the greater trochanter and the lateral ankle during the standing posture.

The X$_{COM}$ was used to calculate the MOS using Equation (\ref{eq:MOS}) \cite{Hof2005}:

\begin{equation}
MOS = BOS - X_{\text{COM}}
\label{eq:MOS}
\end{equation}
In both experiments, only $BOS_A$ was considered, and it was defined by the toe markers. A positive MOS indicates that the X$_{COM}$ remains within the BOS, signifying maintenance of postural stability. Conversely, a negative MOS value reflects X$_{COM}$ displacement beyond the BOS boundary, corresponding to a greater recovery demand or transient dynamic instability \cite{Inagaki2023}.

\begin{figure*}[ht!]
\centering
\includegraphics[width=1\linewidth]{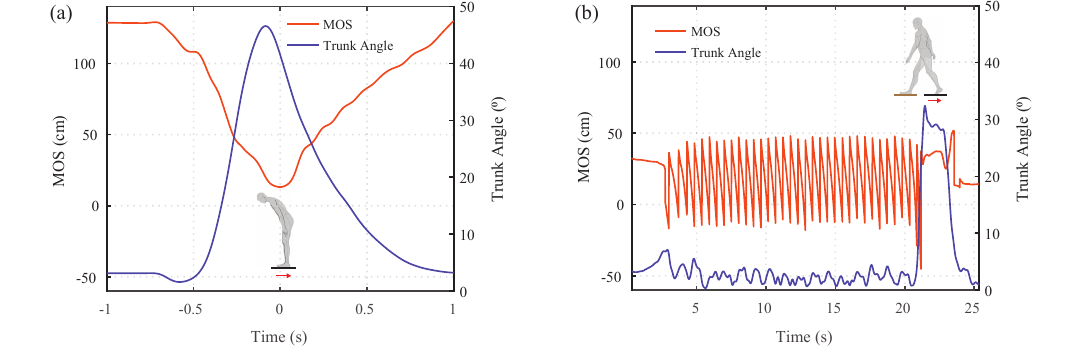}
\caption{Sample results from representative subjects. (a) In the SWP experiment, the subject's trunk angle increases rapidly in response to the perturbation, accompanied by a sharp decrease in MOS. The minimum MOS value at the moment of peak perturbation reflects the subject's lowest stability and is extracted for analysis. (b) In the WWP experiment, after several gait cycles, the subject experiences a sudden forward lean due to the perturbation, resulting in a distinct minimum in MOS before quickly regaining balance. This minimum value is also identified and extracted.}
\label{fig:represent}
\end{figure*}

Motion capture and force plate data were filtered in Vicon Nexus v.2.10 (Vicon Motion Systems Ltd., Oxford, UK) using 4th order, zero-lag, low-pass Butterworth filters with 6 Hz and 20 Hz cut-off frequencies, respectively. The filtered force plate data (GRF) were then downsampled by averaging non-overlapping 10-sample windows in Matlab v.R2024a (The MathWorks, Natick, Massachusetts, USA) \cite{Romanuke2021}, aligning the sampling rate with motion capture data. The 4-point central finite difference method was used to calculate $V_{\text{COM}}$.

%The 4-point central finite difference method demonstrated superior performance over conventional first-order derivative approaches for calculating COM velocity. This method provides enhanced derivative approximation accuracy by reducing truncation errors while simultaneously mitigating the effects of signal noise \cite{LeVeque1998}.

%The mean and standard deviation of the MOS data were calculated for each subject across three stiffness levels in both experiments. Subsequently, a linear mixed-effects model (LME) was applied to the combined dataset across all subjects, accounting for both fixed effects (stiffness levels) and random effects (subject variability) \cite{Galecki2013}. The statistical significance of stiffness level differences identified by the LME was then further verified using two-sided t-tests with a Bonferroni correction at the same significance levels (\(\alpha = 0.05, 0.01, 0.001\)).

For each perturbation trial, the minimum anterior MOS within the perturbation-response window was extracted as the primary outcome measure. The mean and standard deviation of the minimum MOS were calculated for each subject and device condition in the SWP and WWP experiments. To assess group-level effects while accounting for repeated measurements, separate linear mixed-effects models were fitted for SWP and WWP, with minimum MOS as the dependent variable, device condition as a fixed effect, and subject as a random intercept. When a significant main effect of device condition was detected, post hoc pairwise comparisons were performed using two-sided tests with Bonferroni correction. Statistical significance was evaluated at $\alpha = 0.05$, with additional levels indicated at $\alpha = 0.01$ and $\alpha = 0.001$.

%--------------------------------------------------------
\section{Results}

\subsection{Minimum Margin of Stability Extraction}
As shown in Fig. \ref{fig:represent}, the minimum MOS during the perturbation, whether in standing or walking, was identified as the critical value representing the lowest stability point.

%--------------------------------------------------------
\subsection{Individual Results}
\subsubsection{SWP}
All subjects showed an increasing trend in the minimum MOS with higher stiffness levels across all subjects (Table I). By definition, greater MOS values indicate enhanced stability maintenance at the instance of maximal balance loss. All positive mean MOS values across conditions suggest that X$_{COM}$ predominantly remained within the BOS during perturbations, which is consistent with theoretical expectations.

\subsubsection{WWP}
%Each individual subject indicated notable differences in body control strategies during perturbations, and all subjects demonstrated a consistent increasing trend in MOS across stiffness levels, although some of these trends were not obvious. Given that postural control during walking represents a highly chaotic system, coupled with the inherent limitations of the experimental protocol in achieving uniform movements across all subjects, substantial variance in the results is to be expected.
Similar to the results in SWP, the minimum MOS exhibited an overall increasing trend with higher stiffness levels (Table II). However, this trend was less pronounced and less consistent than that observed in SWP. The increased variability in MOS may be attributed to inter-subject differences in postural control strategies during perturbations, to the greater cycle-to-cycle variability inherent to walking, and to limitations in the experimental protocol for enforcing uniform movement patterns across subjects.

\begin{table}[h]
\caption{Minimum MOS During SWP}
\label{tab:standing_mos}
\centering
\begin{tabular}{
p{0.1\columnwidth}
p{0.23\columnwidth}
p{0.23\columnwidth}
p{0.23\columnwidth}
}
\hline
\textbf{Subject} & \textbf{No Device} & \textbf{Device Worn} & \textbf{Device Activated} \\
\hline
S1   & $1.88 \pm 1.09$ & $2.10 \pm 1.59$ & $2.49 \pm 1.98$ \\
S2   & $3.10 \pm 2.57$ & $3.53 \pm 2.72$ & $4.73 \pm 2.88$ \\
S3   & $2.63 \pm 1.55$ & $3.27 \pm 2.47$ & $4.74 \pm 2.02$ \\
S4  & $1.81 \pm 1.25$ & $3.00 \pm 2.00$ & $3.92 \pm 2.00$ \\
S5 & $2.71 \pm 5.70$ & $4.34 \pm 3.72$ & $5.13 \pm 3.65$ \\
\hline
\end{tabular}
\vspace{1ex}
\footnotesize{\\MOS values are reported as mean $\pm$ standard deviation in cm.}
\end{table}

\begin{table}[h]
\caption{Minimum MOS During WWP}
\label{tab:walking_mos}
\centering
\begin{tabular}{
p{0.1\columnwidth}
p{0.23\columnwidth}
p{0.23\columnwidth}
p{0.23\columnwidth}
}
\hline
\textbf{Subject} & \textbf{No Device} & \textbf{Device Worn} & \textbf{Device Activated} \\
\hline
S1 & $-29.51 \pm 7.93$ & $-27.47 \pm 7.13$ & $-24.91 \pm 5.52$ \\
S2 & $-36.15 \pm 7.25$ & $-28.32 \pm 7.17$ & $-28.11 \pm 8.21$ \\
S3 & $-43.14 \pm 8.36$ & $-38.02 \pm 10.32$ & $-35.92 \pm 9.69$ \\
S4 & $-33.76 \pm 6.59$ & $-31.98 \pm 10.10$ & $-29.21 \pm 8.38$ \\
S5 & $-28.09 \pm 3.77$ & $-24.96 \pm 3.32$ & $-23.62 \pm 4.51$ \\
\hline
\end{tabular}
\vspace{1ex}
\footnotesize{\\MOS values are reported as mean $\pm$ standard deviation in cm.}
\end{table}

%\begin{figure*}[!t]
%\centering
%\includegraphics[width=2\columnwidth]{fig/results3.png}
%\caption{Minimum MOS results of each subject. (a) SWP experiment. (b) WWP experiment.}
%\label{fig:finalresult}
%\end{figure*}

%--------------------------------------------------------
\subsection{Group Averaged Results}
After applying the LME model, the statistical results of all subjects are shown below.
\subsubsection{SWP}
As shown in Fig. \ref{fig:LME}(a), there is a clear indication that subjects' stability progressively improved with increasing stiffness levels. The LME analysis showed a significant fixed effect of device condition $(p < 0.001)$. Post hoc pairwise comparisons with Bonferroni correction further demonstrated significant differences among all pairwise comparisons. These findings indicate that the device can effectively improve the subjects' stability at the moment of perturbation at both stiffness levels in the SWP experiment.

\subsubsection{WWP}
\begin{figure}[t!]
\centering
%\vspace{2ex}
\includegraphics[width=1\columnwidth]{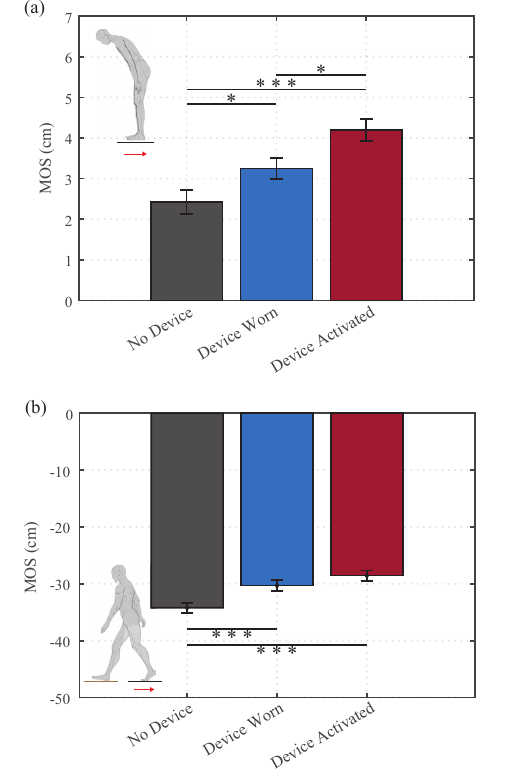}
%\vspace{0.1ex}
\caption{Group averaged results after applying LME model. (a) SWP experiment. (b) WWP experiment.}
\label{fig:LME}
\end{figure}

As shown in Fig. \ref{fig:LME}(b), both the Device Worn (low stiffness) and Device Activated (high stiffness) conditions resulted in significantly higher (less negative) MOS values compared with the No Device condition ($p < 0.001$). No statistically significant difference was observed between the Device Worn and Device Activated conditions. These results indicate that the device, irrespective of stiffness level, meaningfully enhanced postural stability following trip-like perturbations during walking.

%To demonstrate the effect of device stiffness on reactive stability during walking, results from a representative subject are shown in Fig. \ref{fig:WWPMOS} (a). Subject exhibited an increasing trend in minimum MOS across the three conditions, with larger MOS values observed under the Device Activated condition compared to Device Worn and No Device. Despite all values remaining negative—indicating that the XCOM moved beyond the BOS—the reduced magnitude of negative MOS suggests improved recovery capacity with increased stiffness.

%Group-level analysis shown in Fig. \ref{fig:WWPMOS} (b), the Device Activated and Device Worn conditions resulted in significantly higher (less negative) MOS values compared to No Device ($p < 0.001$). The difference between Device Worn and Device Activated was not statistically significant. These results indicate that the device itself assistance meaningfully enhanced postural stability following trip-like perturbations during walking.

% -------------------------------------------------------------
\section{Discussion}
Our lightweight BSD improved postural stability during standing and walking, validating our first hypothesis. MOS increased significantly (Fig. \ref{fig:LME}) during device use (``Device Worn" and ``Device Activated" conditions). Stability improved with increasing stiffness of the device during SWP but not during WWP (Fig. \ref{fig:LME}). Thus, our second hypothesis was validated for standing but not for walking. This may be partly due to large inter-subject variability in the outcome measures, which may have reduced sensitivity to stiffness-dependent effects during WWP despite standardized perturbation magnitudes, device stiffness settings, and task constraints. This variability likely reflects individual differences in body mechanics and neuromuscular balance-recovery strategies. Future investigations may benefit from implementing personalized calibration protocols for subject-specific neuromechanical response patterns. Another possible reason is that the maximum trunk flexion elicited during WWP (30 ± 3°) was lower than that of SWP (35 ± 3°). The limited difference in the forces generated by the device between the ``Device Worn" and ``Device Activated" conditions (Fig. \ref{fig:device}(d)) at smaller trunk angles may have diminished the difference between the two device conditions in WWP. Overall, these findings suggest that the BSD can enhance balance stability, particularly during standing perturbations, while highlighting the need for subject-specific tuning and greater assistive force differentiation to maximize benefits during walking. 

%Second, the treadmill's electromechanical limitations inherently restricted perturbation control fidelity. While acceleration and velocity parameters were precisely programmed, the inability to implement time-based motion profiles necessitated manual calculation of perturbation durations. Furthermore, substantial motor inertia introduced minor but non-negligible variations in instantaneous acceleration profiles across trials. These electromechanical tolerances introduced systematic noise into temporal perturbation characteristics.

Despite promising improvements in MOS, our study has several limitations. First, our sample size was limited to five healthy young adults. Further, we evaluated stability using only MOS. Though MOS is a well-established stability quantification metric, recent literature shows that the whole-body angular momentum (WBAM) may be another promising metric \cite{leestma2023linking,nazifi2020angular}. In the future, we plan to extend our sample size, include elderly subjects, and evaluate multiple stability metrics to strengthen our conclusions. Another limitation in the SWP experiment was that we tested perturbation magnitudes that did not elicit stepping. In real-world scenarios, some perturbations may require stepping to maintain balance. Future experiments should also evaluate our device's efficacy under perturbations that require stepping.

Regarding our experimental setup, one limitation stems from the trade-off between practicality and biomechanical accuracy in the simplified marker set. Compared to the full-body Plug-in Gait model incorporating anthropometric measurements for skeletal modeling, our simplified configuration relied solely on marker geometry for COM estimation. Although Vicon's official documentation acknowledges inherent COM approximation errors even in full configurations, particularly during non-gait motions like stooping or squatting \cite{Vicon2023}, the simplified approach likely amplified positional uncertainties during dynamic instability events. Further, fundamental differences in locomotor control strategies between treadmill and overground walking introduced ecological validity concerns \cite{Tielke2019,Alton1998}. Treadmill-based perturbation protocols artificially decouple propulsion mechanics from spatial navigation.

%\vspace{-0.07ex}
\section{Conclusion}
This study evaluated the effectiveness of a BSD in improving human stability under perturbation conditions. Two distinct experimental paradigms, Standing with Perturbations (SWP) and Walking with Perturbations (WWP), were employed to assess how different device stiffness levels influence the margin of stability (MOS). Results indicated an increasing trend in stability with greater device stiffness, as evidenced by higher mean MOS values during moments of maximal balance loss. Statistical analyses confirmed that both ``Device Worn" and ``Device Activated" conditions produced significant improvements in MOS compared to ``No Device" in both SWP and WWP experiments. However, no significant difference was observed between ``Device Worn" and ``Device Activated" in the WWP condition. This outcome may be attributed partly to inter-subject variability, reflecting inherent differences in individual biomechanical and neuromuscular control strategies, and partly to a limited difference between the ``Device Worn" and ``Device Activated" conditions at small trunk flexion angles.

Several limitations were identified, including the small sample size of five healthy young adults, reliance on MOS as the sole stability metric, use of standing perturbations that did not elicit stepping responses, mechanical constraints of the treadmill system, reduced COM estimation accuracy associated with the simplified marker set, and fundamental biomechanical differences between treadmill and overground walking. Future research should address these limitations by increasing the sample size, including elderly subjects, evaluating additional stability metrics such as whole-body angular momentum, and testing perturbations that require stepping to maintain balance. Additionally, future studies could employ the improved BSD currently being developed by Khatavkar’s team \cite{khatavkar2026soft}, which integrates an electroadhesive clutch system. This advanced device allows for higher resolution and accuracy in stiffness control, potentially offering enhanced support during unexpected trip perturbations. Despite these limitations, the present study contributes valuable insights into the potential of wearable BSDs for improving human stability in response to perturbations.

% REFERENCE------------------------------------------------------

\end{document}